\documentclass[letterpaper]{article} 
\usepackage{aaai25}  
\usepackage{times}  
\usepackage{helvet}  
\usepackage{courier}  
\usepackage[hyphens]{url}  
\usepackage{graphicx} 
\usepackage{natbib}  
\usepackage{caption} 
\usepackage{algorithm}
\usepackage{algorithmic}
\usepackage{enumitem}
\usepackage{xcolor}
\usepackage{amsmath}
\usepackage{amssymb}

\usepackage{xcite}
\usepackage{amsfonts}
\usepackage{textcomp}

\usepackage{multirow}
\usepackage{bigstrut}
\usepackage{makecell}
\usepackage{pifont}
\usepackage{subfloat}
\usepackage{diagbox} 
\usepackage{placeins}
\usepackage{boldline}
\usepackage{booktabs}
\usepackage{subcaption}
\usepackage{framed}
\usepackage{color}

\definecolor{shadecolor}{gray}{1} 

\usepackage[inkscapelatex=false]{svg}

\usepackage{cleveref}
\crefname{proposition}{Proposition}{Propositions}

\usepackage{newfloat}
\usepackage{listings}
\DeclareCaptionStyle{ruled}{labelfont=normalfont,labelsep=colon,strut=off} 
\floatstyle{ruled}
\newfloat{listing}{tb}{lst}{}
\floatname{listing}{Listing}
\title{GDiffRetro: Retrosynthesis Prediction with Dual Graph Enhanced\\ Molecular Representation and Diffusion Generation}
\author{
    Shengyin Sun\textsuperscript{\rm 1}\thanks{These authors contributed equally.}$^\dag$,
    Wenhao Yu\textsuperscript{\rm 2}$^*$\thanks{Work done as interns at Huawei ACS Lab.}, 
    Yuxiang Ren\textsuperscript{\rm 3}\thanks{Corresponding author.}, 
    Weitao Du\textsuperscript{\rm 4}, 
    Liwei Liu\textsuperscript{\rm 3},
    Xuecang Zhang\textsuperscript{\rm 3},\\
    Ying Hu\textsuperscript{\rm 5},
    Chen Ma\textsuperscript{\rm 1}
}
\affiliations{
    \textsuperscript{\rm 1}Department of Computer Science, City University of Hong Kong \\
    \textsuperscript{\rm 2}Department of Computer Science and Engineering, The Chinese University of Hong Kong \\
    \textsuperscript{\rm 3}Advance Computing and Storage Lab, Huawei Technologies \\
    \textsuperscript{\rm 4}Academy of Mathematics and Systems Science, Chinese Academy of Sciences \\
    \textsuperscript{\rm 5}Department of Electronic Engineering, Tsinghua University
    
    shengysun4-c@my.cityu.edu.hk, whyu24@cse.cuhk.edu.hk, \{renyuxiang1, liuliwei5, zhangxuecang\}@huawei.com, duweitao@amss.ac.cn, yinghu\_yh@mail.tsinghua.edu.cn, chenma@cityu.edu.hk
}

\usepackage{bibentry}

\begin{document}

\maketitle

\begin{abstract}
Retrosynthesis prediction focuses on identifying reactants capable of synthesizing a target product. Typically, the retrosynthesis prediction involves two phases: Reaction Center Identification and Reactant Generation. However, we argue that most existing methods suffer from two limitations in the two phases: (i) Existing models do not adequately capture the ``face'' information in molecular graphs for the reaction center identification. (ii) Current approaches for the reactant generation predominantly use sequence generation in a 2D space, which lacks versatility in generating reasonable distributions for completed reactive groups and overlooks molecules' inherent 3D properties. To overcome the above limitations, we propose GDiffRetro. For the reaction center identification, GDiffRetro uniquely integrates the original graph with its corresponding dual graph to represent molecular structures, which helps guide the model to focus more on the faces in the graph. For the reactant generation, GDiffRetro employs a conditional diffusion model in 3D to further transform the obtained synthon into a complete reactant. Our experimental findings reveal that GDiffRetro outperforms state-of-the-art semi-template models across various evaluative metrics.
\end{abstract}

%

\section{Introduction}
\label{ssy0119:introduction}
The retrosynthesis task aims to find a set of reactants capable of synthesizing a given product, which is a one-to-many problem. Even for experienced chemists, addressing such a one-to-many problem is still extremely challenging. Recently, benefitting from the rapid advancement of deep learning (DL), researchers have resorted to DL models to design efficient methods for retrosynthesis tasks. Existing DL-based methods~\cite{somnath2021learning} can be divided into template-based, template-free, and semi-template methods. Template-based methods rely on predefined templates (extracted from a large-scale chemical database). For example, the GLN \cite{dai2019retrosynthesis} treats chemical knowledge of reaction templates as logical rules, and then models the joint probability between rules and reactants. As template-based methods are constrained by predefined templates, template-free approaches are proposed. For example, Chemformer~\cite{irwin2022chemformer} formulates the retrosynthesis prediction as a translation task, where the product SMILES (strings describing molecular compositions) and the set of reactant SMILES serve as the ``source language strings'' and ``target language strings'', respectively. Template-free methods typically generate reactant SMILES by sequentially outputting individual symbols, which makes their predictions limited in diversity.

To alleviate issues present in both template-based and template-free methods, the semi-template framework has recently been adopted, which does not utilize reaction templates and has good interpretability. It predicts the final reactants through the intermediates (synthons) in two steps: first identifying the reaction center to form synthons, then completing the synthons into reactants. For instance, G2Gs~\cite{shi2020graph} first employs the Relational Graph Convolutional Network (RGCN) \cite{arxiv:SchlichtkrullKipf17} for reaction center identification, and then generates products through the variational graph translation. 

Although existing semi-template methods have achieved success in some scenarios, we argue that there are still avenues to improve. First, classical methods~\cite{shi2020graph,dai2019retrosynthesis} solely focus on the nodes within the molecular graph, neglecting the features associated with faces (edges divide the entire plane into a set of regions, called faces) in the graph. The features of faces play a crucial role in the reaction center identification. For example, in a benzene ring, all carbon atoms reside on one face, and the bonds connecting these carbons exhibit high stability, making them less likely to serve as reaction centers. Secondly, existing methods generate reactants based on 2D graphs, ignoring the 3D structural information of molecules to some extent. 

To address the above shortcomings, we propose the retrosynthesis prediction with dual graph-enhanced molecular representation and diffusion generation (\textbf{GDiffRetro}). We first introduce dual graphs to the reaction center identification. The dual graph is a way to describe a graph from the perspective of its faces. In the dual graph, each node corresponds to a face in the original graph. Our primary motivation for introducing the dual graph is to integrate the face information into the node representations, enabling the model to focus on faces within molecular graphs, such as distinguishing whether different nodes are on the same faces. Considering the powerful capability of generative models in capturing realistic patterns \cite{DBLP:conf/iccv/ZhangRao23, SunRen23}, we then employ the 3D diffusion model to generate final reactants. Specifically, we generate reactants conditioned on the synthons obtained from the reaction center identification stage, and then conduct the 3D diffusion process to preserve the reactants' inherent structural properties. Our contributions are summarized below:
\begin{itemize}[leftmargin=*]
\item{To better extract information from molecular graphs, we introduce dual graphs to guide the model in focusing on the faces in the molecular structure, which enables the model to precisely identify reaction centers.}
\item{To better transform intermediates obtained from the stage of reaction center identification into reactants, we make the first attempt at resorting to the conditional diffusion model in the semi-template retrosynthesis prediction.}
\end{itemize}

\section{Related Work}
\textbf{Retrosynthesis Prediction}.
 Some significant template-based works include GLN \cite{dai2019retrosynthesis}, LocalRetro \cite{chen2021deep}, and Dual-TB \cite{sun2020energy}. To overcome the constraints of external knowledge in template-based methods, template-free methods have been developed. Key examples in the template-free category include Chemformer \cite{irwin2022chemformer}, RetroBridge  \cite{DBLP:conf/iclr/IgashovSchneuing24}, and Dual-TF \cite{sun2020energy}. Considering that template-free methods lack interpretability, semi-template methods have emerged. Notable works in the semi-template category include MEGAN \cite{sacha2021molecule}, G2Gs \cite{shi2020graph}, and RetroDiff \cite{WangSong24}.
 
\vspace{0.1cm}
\noindent\textbf{Molecular Generation}.
The molecular generation is closely related to generative models~\cite{ZhangLin24,GuoLiu24,ZhangRen24}. For example, You et al. modeled the molecular generation as a sequential decision process on graphs \cite{YouLiu18}. Recent works introduced diffusion models to molecular data. GeoDiff \cite{xu2022geodiff} and ConfGF \cite{shi2021learning} condition the model on the adjacency matrix of the molecular, enabling them to optimize torsion angles between atoms. The equivariant diffusion model \cite{hoogeboom2022equivariant} generates 3D molecules from scratch, conditioned on predefined scalar properties. Another noteworthy work is DiffLinker \cite{igashov2022equivariant}, an 3D diffusion model for designing linkers. Other models include LatentDiff \cite{DBLP:conf/logs/CongYan24} for protein design and AbX \cite{DBLP:conf/icml/ZhuRen24} for antibody design.

\section{Methodology}
\label{ssy1210:method}
\begin{figure*}
    \centering
    \includegraphics[scale=0.36]{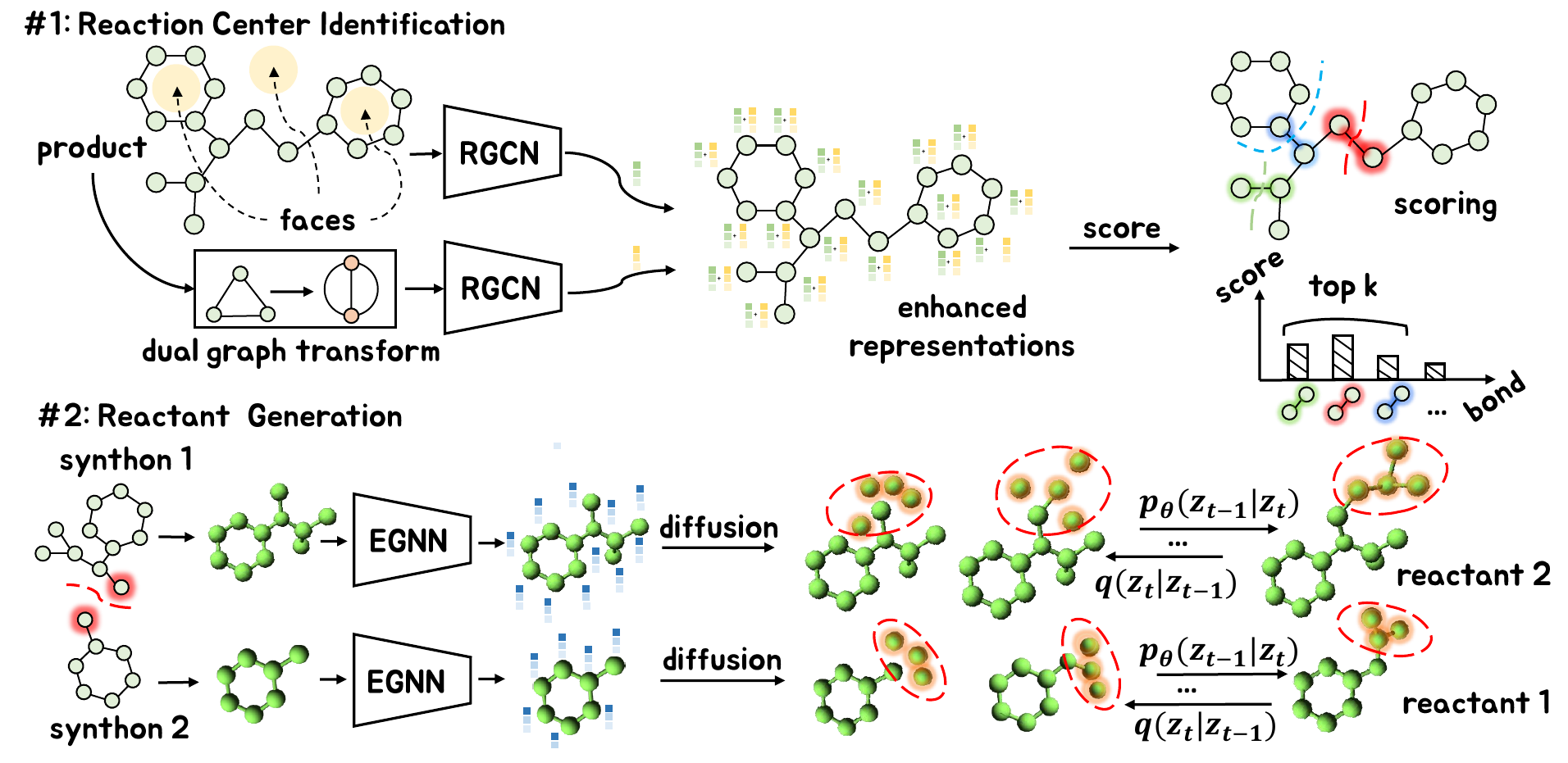}
    \caption{The framework of GDiffRetro. In the stage \#1, we utilize the dual graph to enhance the representations. In the stage \#2, we employ the 3D diffusion model (conditioned on the obtained synthon) to convert synthons into reactants.}
    \label{ssy1210:framework}
\end{figure*}

\subsection{Graph-based Reaction Center Identification} 
\label{ssy0722:problem_formulation}
A molecule $\mathcal{M}$ containing $n$ atoms and $q$ types of chemical bonds can be written as $\mathcal{M}=\left\{{\mathbf{A}}, \mathbf{X}\right\}$, where $\mathbf{X}\in\mathbb{R}^{n\times d}$ is a $d$-dimensional node feature matrix and $\mathbf{A}\in\mathbb{R}^{n\times n\times q}$ is an adjacency matrix ($A_{i,j,k}=1$ if there exists a bond of type $k$ between atom $i$ and atom $j$). Based on the above formulation, a chemical reaction can be described as a pair of sets $\left(\mathcal{G}^{\rm{r}}, \mathcal{G}^{\rm{p}}\right)$, where $\mathcal{G}^{\rm{r}}=\left\{\mathcal{M}_{i}^{\rm{r}}\right\}\vert_{i=1}^{l}$ is a set containing $l$ reactants and $\mathcal{G}^{\rm{p}}=\{\mathcal{M}_{j}^{\rm{p}}\}\vert_{j=1}^{m}$ is a set containing $m$ products. Following previous work, we focus only on standard single-output chemical reactions, i.e., $\left|\mathcal{G}^{\rm{p}}\right|=1$. For a single-output reaction $\left(\left\{\mathcal{M}_{i}^{\rm{r}}\right\}\vert_{i=1}^{l}, \mathcal{M}^{\rm{p}}\right)$, the goal of retrosynthesis is to predict the set of reactants $\left\{\mathcal{M}_{i}^{\rm{r}}\right\}\vert_{i=1}^{l}$ corresponding to the given product $\mathcal{M}^{\rm{p}}$. The overview of our solution for the retrosynthesis task is shown in Fig. \ref{ssy1210:framework}. We first conduct reaction center identification to partition the products into synthons (subgraphs of the product molecule, often not valid molecules). Then, we utilize a diffusion model to generate reactants based on the previously obtained synthons. Notations and chemical terms are summarized in Section \ref{ssy0724:notations} and  Section \ref{ssy0724:terms} of the supplementary material, respectively. The supplementary material is in arXiv version of the paper.

Given embeddings of two atoms in the product, the prediction model for reaction center identification is required to output a score, i.e., the probability of a reaction center existing between these two atoms. The higher the probability, the more it indicates that the product needs to break the bond between these two atoms to generate synthons. 

Considering the heterogeneity of the molecular graph, we use the RGCN to encode atoms in the given product $\mathcal{M}^{\rm{p}}=\left\{\mathbf{A}^{\rm{p}}, \mathbf{X}^{\rm{p}}\right\}$. \textcolor{black}{The node $i$'s representation is updated as follows (starting with node $i$'s initial feature $\mathbf{h}_{i}^{0} = \mathbf{X}^{\rm p}\left[i,:\right]$):}
\begin{equation}
    \label{ssy0119:node_embedding}
        \left\{\mathbf{h}_{i}^{l}\right\}\vert_{l=1}^{L} = \sigma\Big(\sum_{r\in\mathcal{R}}\sum_{j\in\mathcal{N}_{i}^{r}}\mathbf{W}_{r}^{l-1}\mathbf{h}^{l-1}_{j}+\mathbf{W}^{l-1}_{0}\mathbf{h}_{i}^{l-1}\Big),
\end{equation}
where $\mathcal{R}$ is the set of all edge types (chemical bonds), $\mathcal{N}_{i}^{r}$ is the set of neighbors of node $i$ under relation $r$ (can be obtained through $\mathbf{A}^{\rm{p}}[:,:,r]$), $\sigma(\cdot)$ is an activation function, $\mathbf{W}_{r}$ is the learnable weight matrix corresponding to the edge type $r$, and $\mathbf{W}_{0}$ is the learnable weight matrix for the self-loop edge. An RGCN with $L$ layers can only aggregate information from nodes within $L$ hops, while the reactivity of a reaction center may also be related to more distant nodes. Therefore, we also compute the graph-level embedding (by applying the $\rm{Readout}(\cdot)$ function proposed in ~\cite{DBLP:conf/iclr/VelickovicFedus19}) to introduce the influence of remote atoms, i.e.,
\begin{equation}
    \mathbf{h}_{\mathcal{M}^{\rm p}} = \rm{Readout}(\mathbf{H}^{L}),
\end{equation}
where $\mathbf{H}^{L}$ is a node embedding matrix constructed from $\mathbf{h}_{i}^{L}$.

\subsection{Dual Graph Enhanced Representation}
\label{ssy0722:Dual_Enhanced}
\begin{figure}[h]
    \centering
    \includegraphics[width=0.338\textwidth]{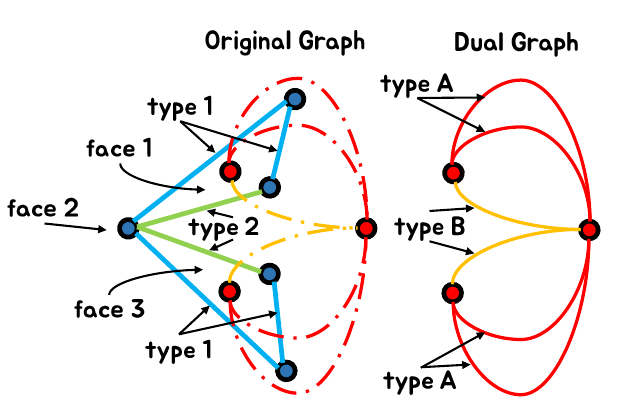}
    \caption{An example of dual graph construction. Each node in the dual graph corresponds to a face in the original graph, and the type of each edge in the dual graph depends on the type of the edge it crosses in the original graph (Type 1 $\Leftrightarrow$ Type A, Type 2 $\Leftrightarrow$ Type B). 
    More details about dual graphs and faces are in Section ~\ref{ssy0724:dual_graph_explain} (supplementary material).}
    \label{ssy0119:dual graph-disp}
\end{figure}

Encoding strategies in Section \ref{ssy0722:problem_formulation} is designed from the perspective of the nodes, while neglecting the faces in the molecular graph. Faces in the molecular graph are equally important. For example, carbon atoms in a benzene ring are all in one face, and the bonds between these carbons are very stable, making them unlikely to become reaction centers. To make up this shortcoming, we introduce the dual graph $\mathcal{D}^{\rm{p}}=\left\{\mathbf{A}^{\rm{p}}_{\rm d}, \mathbf{X}^{\rm{p}}_{\rm{d}}\right\}$ of $\mathcal{M}^{\rm{p}}$, which is constructed as follows:
\begin{enumerate}[left=0pt, label=$\bullet$]
    \item {\textbf{Topological structure construction}. Given an original planar graph $\mathcal{M}^{\rm{p}}$, the dual graph $\mathcal{D}^{\rm{p}}$ is a graph that has a node for each face of $\mathcal{M}^{\rm{p}}$. Additionally, $\mathcal{D}^{\rm{p}}$ has an edge connecting two nodes if the corresponding faces in $\mathcal{M}^{\rm{p}}$ are separated by an edge in the $\mathcal{M}^{\rm{p}}$. The type of an edge in $\mathcal{D}^{\rm{p}}$ corresponds to the type of the edge it crosses in $\mathcal{M}^{\rm{p}}$. As shown in Fig. \ref{ssy0119:dual graph-disp}, five nodes in the original graph divide the entire space into three parts. Consequently, its corresponding dual graph contains three nodes, with each node representing a face. Furthermore, the original graph contains two types of edges (denoted as blue and green, respectively). Similarly, its dual graph also comprises two types of edges. Specifically, edges crossing the blue edges in the original graph belong to one type (marked as red), whereas edges crossing the green edges in the original graph belong to another type (marked as orange).}
    \item {\textbf{Node feature construction}. The feature of a node in the dual graph depends on the surface where the node is located. Formally, the feature of node $i$ in $\mathcal{D}^{\rm{p}}$ is:
    \begin{equation}
    \fontsize{9.3pt}{12pt}\selectfont
        \mathbf{X}^{\rm{p}}_{\rm{d}}[i,:] = \frac{1}{\vert\mathcal{S}_{i}\vert}\sum_{j\in\mathcal{S}_{i}}\mathbf{X}^{\rm{p}}[j,:],
    \end{equation}
    where $\mathcal{S}_{i}$ is the set of nodes in $\mathcal{M}^{\rm{p}}$ on the face where node $i$ in $\mathcal{D}^{\rm{p}}$ is located.}
\end{enumerate}

After obtaining $\mathcal{D}^{\rm{p}}$, we encode the nodes in the dual graph in the same way as before (through an L-layer RGCN), i.e., 
\begin{equation}
    \mathbf{D}^{L} = \rm{RGCN}(\mathcal{D}^{\rm{p}}),
\end{equation}
where each row in $\mathbf{D}^{L}$ is the final embedding of a node in $\mathcal{D}^{\rm{p}}$. Combining $\mathcal{M}^{\rm{p}}$ and $\mathcal{D}^{\rm{p}}$, the final embedding of node $i$ in the original molecular graph can be expressed as:
\begin{equation}
\label{ssy0119:enhanced_node_embedding}
    \mathbf{m}_{i} = \mathbf{H}^{L}[i,:]~\|~\sum_{j\in\mathcal{F}_{i}}\mathbf{D}^{L}[j,:]~\|~\mathbf{h}_{\mathcal{M}^{\rm{p}}},
\end{equation}
where $\mathcal{F}_{i}$ is the set of nodes in $\mathcal{D}^{\rm{p}}$ on the face where node $i$ in $\mathcal{M}^{\rm{p}}$ is located. In order to estimate the reactivity probability between a pair of nodes $i$ and $j$, we formulate the edge embedding as follows:
\begin{equation}
    \mathbf{e}_{ij} = \mathbf{m}_{i}~\|~\mathbf{m}_{j}~\|~\mathbf{A}^{\rm p}[i,j,:].
\end{equation}
Then, the reactivity score can be calculated as $s_{ij} = \rm{Sigmoid}\left(\phi(\mathbf{e}_{ij})\right)$, where $\phi(\cdot)$ is a network for converting edge embeddings to scalar scores. For training, the proposed module is optimized by maximizing the following loss:
\begin{equation}
\fontsize{9.3pt}{12pt}\selectfont
    \mathcal{L}^{\rm(1)} = -\mathbb{E}_{\mathcal{P}_{\rm{r}}}\Big[\sum_{i}\sum_{j\neq i}\lambda Y_{ij}{\rm{log}}(s_{ij})+(1-Y_{ij}){\rm{log}}(1-s_{ij})\Big],
\end{equation}
where $\mathcal{P}_{\rm{r}}$ is the set of all chemical reactions in the training data, $Y_{ij}$ is the true label indicating whether a reaction center exists between atoms $i$ and $j$, and $\lambda$ is a hyper-parameter for alleviating class imbalance issues.

In Fig. \ref{ssy0119:dual graph-disp}, each node in the dual graph represents a face in the original graph. When we input the dual graph into the RGCN, the outputted node representations of the dual graph are actually the face representations of the original graph. The obtained face representations are then integrated into the node representations of the original graph through Eq. \ref{ssy0119:enhanced_node_embedding}. 
In this way, the face information within the original graph can
be effectively used. When two nodes in the original graph exist 
within the same face, their corresponding face representations included in the Eq. \ref{ssy0119:enhanced_node_embedding} will be the same (i.e., they have the same face information), which aligns with intuition.

\subsection{Conditional Diffusion Reactant Generation}
\noindent In Fig. \ref{ssy1210:framework}, the conditional diffusion generation involves two processes: \textit{i)} A forward process corrupts the structure and features of a synthon by adding Gaussian noise. \textit{ii)} A reverse process learns the denoise process and outputs a reactant. 

\vspace{0.1cm}
\noindent\textbf{$\bullet$ Forward process.} An atom $\mathbf{s}$ can be represented by a 3D coordinates $\mathbf{u}^{(\rm x)}\in\mathbb{R}^{3}$ and $d$-dimensional features $\mathbf{u}^{(\rm h)}\in\mathbb{R}^{\rm{d}}$, i.e., $\mathbf{s}=[\mathbf{u}^{(\rm x)},\mathbf{u}^{(\rm h)}]$. Setting $\mathbf{z}_{0}=\mathbf{s}$ as the initial state and parameterize a fixed noise process as: $q\left(\mathbf{z}_{t}\vert\mathbf{z}_{0}\right)=\mathcal{N}\left(\mathbf{z}_{t};\alpha_{t}\mathbf{z}_{0},\sigma_{t}^{2}\mathbf{I}\right), t=1,\cdots,T,$ where $\mathbf{z}_t$ is a latent noised representation, $\mathcal{N}(\cdot)$ denotes a Gaussian distribution, $\mathbf{I}$ is an identity matrix, $\alpha_{t}$ controls the proportion of the original input to be retained, and $\sigma_{t}^{2}$ controls the intensity of added Gaussian noise. Inspired by \cite{DBLP:conf/iclr/song2021scorebased}, we adopt a variance-preserving noise adding process, i.e.,
$\alpha_t=\sqrt{1-\sigma_{t}^{2}}$. In addition, a numerically stable polynomial noise schedule \cite{hoogeboom2022equivariant} is applied, i.e., $\alpha_t = (1-2s)\left[1-(\frac{t-1}{T-1})^2\right]$, where $s$ is set to $10^{-5}$ to ensure numerical stability.

\vspace{0.1cm}
\noindent\textbf{$\bullet$ Reverse process.} The reverse process takes $\mathbf{x}_{T}$ as the starting point and attempts to learn a network with $\boldsymbol{\theta}$ as a trainable parameter for denoising, as follows:
\begin{equation}
\fontsize{9.3pt}{12pt}\selectfont
\label{ssy0119:p_theta}
    p_{\boldsymbol{\theta}}(\mathbf{z}_{t-1}\vert\mathbf{z}_{t})=\mathcal{N}(\mathbf{z}_{t-1};\boldsymbol{\mu}_{\boldsymbol{\theta}}(\mathbf{z}_{t}, t), \mathbf{R}_{\boldsymbol{\theta}}(\mathbf{z}_{t}, t)),
\end{equation}
where $\boldsymbol{\mu}_{\boldsymbol{\theta}}(\mathbf{z}_{t}, t)$ and $\mathbf{R}_{\boldsymbol{\theta}}(\mathbf{z}_{t}, t)$ are obtained from a network parameterized by $\boldsymbol{\theta}$. 

\vspace{0.1cm}
\noindent\textbf{$\bullet$ Optimization process.}
To maximize the likelihood of observed data, we optimize the variational lower bound, i.e.,
\begin{equation}
\fontsize{10pt}{12pt}\selectfont
    \begin{aligned}
-&\log p_{\boldsymbol{\theta}}(\mathbf{z}_0) =-\log\int p_{\boldsymbol{\theta}}(\mathbf{z}_{0:T})\mathrm{d}\mathbf{z}_{1:T}\\
& \le \textstyle\sum_{t>1}^{T}\underbrace{\mathbb{E}_{q(\mathbf{z}_t|\mathbf{z}_0)}\left[D_{\text{KL}}(q(\mathbf{z}_{t-1}|\mathbf{z}_t,\mathbf{z}_0)\parallel p_{\boldsymbol{\theta}}(\mathbf{z}_{t-1}|\mathbf{z}_t))\right]}_{\small(\text{diffusion loss $\mathcal{L}_{t}$})}\\
& + \underbrace{D_{\text{KL}}(q(\mathbf{z}_T|\mathbf{z}_0)\parallel p(\mathbf{z}_T))}_{\small(\text{prior loss $\mathcal{L}_{\rm{p}}$})} -\underbrace{\mathbb{E}_{q(\mathbf{z}_1|\mathbf{z}_0)}\left[\log p_{\boldsymbol{\theta}}(\mathbf{z}_0|\mathbf{z}_1)\right]}_{\small(\text{reconstruction loss $\mathcal{L}_{\rm{1}}$})},
\end{aligned} 
\end{equation}
where $p_{\boldsymbol{\theta}}(\mathbf{z}_{0:T})=p_{\boldsymbol{\theta}}(\mathbf{z}_{0},\mathbf{z}_{1},\cdots,\mathbf{z}_{T})$, $\mathrm{d}\mathbf{z}_{1:T}=\mathrm{d}\mathbf{z}_{1}\mathrm{d}\mathbf{z}_{2}\cdots\mathrm{d}\mathbf{z}_{T}$, and $D_{\text{KL}}(\cdot)$ is the KL divergence. The $\mathcal{L}_{{t}}$ encourages approximating the $q(\mathbf{z}_{t-1}|\mathbf{z}_t,\mathbf{z}_0)$ through a distribution $p_{\boldsymbol{\theta}}(\mathbf{z}_{t-1}|\mathbf{z}_t)$ associated with a network. The closed form of $q(\mathbf{z}_{t-1}|\mathbf{z}_t,\mathbf{z}_0)$ can be expressed as:
\begin{equation}
\label{ssy0119:q_closed_form}
    q(\mathbf{z}_{t-1}|\mathbf{z}_t,\mathbf{z}_0)=\mathcal{N}(\mathbf{z}_{t-1};{{\boldsymbol{\mu}}_{{q}}}(\mathbf{z}_t, \mathbf{z}_0, t),\sigma^2_{{q}}(t)\mathbf{I}),
\end{equation}
where
\begin{equation}
\label{ssy0119:mu_q}
{\boldsymbol{\mu}}_{q}(\mathbf{z}_t,\mathbf{z}_0,t) =\frac{\alpha_t\sigma_{t-1}^2}{\alpha_{t-1}\sigma_{t}^2}\mathbf{z}_t+\frac{\alpha_{t-1}^2\sigma_{t}^2-\alpha_{t}^2\sigma_{t-1}^2}{\alpha_{t-1}\sigma_{t}^2}\mathbf{z}_0,
\end{equation}
and
$\sigma^2_{q}(t) = \sigma_{t-1}^2-\frac{\alpha_{t}^2\sigma_{t-1}^4}{\alpha_{t-1}^2\sigma_t^{2}}$. Substituting Eq. \ref{ssy0119:p_theta} and Eq. \ref{ssy0119:q_closed_form} into the diffusion loss at the time step $t$ yields:
\begin{equation}
\fontsize{9.3pt}{12pt}\selectfont
    \label{ssy0119:diffusion_loss_t}
    \begin{aligned}
        \mathcal{L}_{t} & = \mathbb{E}_{q(\mathbf{z}_t|\mathbf{z}_0)}\left[D_{\text{KL}}(q(\mathbf{z}_{t-1}|\mathbf{z}_t,\mathbf{z}_0)\parallel p_{\boldsymbol{\theta}}(\mathbf{z}_{t-1}|\mathbf{z}_t))\right] \\
        & = \mathbb{E}_{q(\mathbf{z}_t|\mathbf{z}_0)}\left[\dfrac{1}{2\sigma^2_{q}(t)}\left[\parallel\boldsymbol{\mu}_{\boldsymbol{\theta}}(\mathbf{z}_t,t)-{\boldsymbol{\mu}_{q}}(\mathbf{z}_t,\mathbf{z}_0,t)\parallel_2^2\right]\right],
    \end{aligned}
\end{equation}
where ${\boldsymbol{\mu}}_{{\boldsymbol{\theta}}}(\mathbf{z}_t,t)$ can be expressed according to the Eq. \ref{ssy0119:mu_q}:
\begin{equation}
\fontsize{9.3pt}{12pt}\selectfont
    \label{ssy0119:mu_theta_close_form}
    {\boldsymbol{\mu}}_{\boldsymbol{\theta}}(\mathbf{z}_t,t) =\frac{\alpha_t\sigma_{t-1}^2}{\alpha_{t-1}\sigma_{t}^2}\mathbf{z}_t+\frac{\alpha_{t-1}^2\sigma_{t}^2-\alpha_{t}^2\sigma_{t-1}^2}{\alpha_{t-1}\sigma_{t}^2}\mathbf{\hat z}_{\boldsymbol{\theta}}(\mathbf{z}_t,t),
\end{equation}
with $\mathbf{\hat z}_{\boldsymbol{\theta}}(\mathbf{z}_t,t)$ as the predicted initial state $\mathbf{z}_0$ (output by a network). Plugging Eq. \ref{ssy0119:mu_q} and Eq. \ref{ssy0119:mu_theta_close_form} into Eq. \ref{ssy0119:diffusion_loss_t} results in
\begin{equation}
\label{ssy0119:L_dt_x}
\fontsize{9.3pt}{12pt}\selectfont
    \begin{aligned}
        \mathcal{L}_{t} = \mathbb{E}_{q(\mathbf{z}_t|\mathbf{z}_0)}\left[\dfrac{1}{2}\left(\frac{\alpha_{t-1}^2}{\sigma_{t-1}^2}-\frac{\alpha_{t}^2}{\sigma_{t}^2}\right)\parallel\hat{\mathbf{z}}_{\boldsymbol{\theta}}(\mathbf{z}_t,t)-\mathbf{z}_0\parallel_2^2\right].
    \end{aligned}
\end{equation}
Considering that $\mathbf{z}_{t}$ can be reparameterized as $\mathbf{z}_{t}=\alpha_t\mathbf{z}_{0}+\sigma_t\boldsymbol\epsilon$, where $\boldsymbol{\epsilon}\sim\mathcal{N}(\mathbf{0},\mathbf{I})$, \cite{ho2020denoising} suggests that using a network to predict $\boldsymbol{\epsilon}$ instead of $\mathbf{z}_{0}$ will lead to a better result, i.e., $\mathcal{L}_{t}$ can be simplifies to:
\begin{equation}
\fontsize{9.3pt}{12pt}\selectfont
    \label{ssy0119:simple_Lt}
    \mathcal{L}_{t} = \mathbb{E}_{\epsilon\sim\mathcal{N}(\mathbf{0}, \mathbf{I})}\left[\dfrac{1}{2}\left(\frac{\alpha_{t-1}^2\sigma_{t}^2}{\alpha_{t}^2\sigma_{t-1}^2}-1\right)\parallel\hat{\boldsymbol{\epsilon}}_{\boldsymbol{\theta}}(\mathbf{z}_t,t)-\boldsymbol{\epsilon}\parallel_2^2\right].
\end{equation}
According to ~\cite{hoogeboom2022equivariant}, both $\mathcal{L}_{\rm p}$ and $\mathcal{L}_{\rm 0}$ are close to 0 (due to $\alpha_{T}=0$, $\alpha_{1}\approx1$ , and $\mathbf{z}_{0}$ is discrete). Furthermore, Ho \emph{et al.} \cite{ho2020denoising} found that removing the weight in Eq. \ref{ssy0119:simple_Lt} is conducive to improving sample quality. Therefore, an unweighted version of the final loss $\mathcal{L}^{(2)}$ used in the reactant generation phase is:
\begin{equation}
    \label{ssy0119:final_simple_Lt}
    \fontsize{9.3pt}{12pt}\selectfont
    \mathcal{L}^{(2)} = \mathbb{E}_{\epsilon\sim\mathcal{N}(\mathbf{0}, \mathbf{I}),t\sim\mathcal{U}(1,T)}\left[\parallel\hat{\boldsymbol{\epsilon}}_{\boldsymbol{\theta}}(\mathbf{z}_t,t)-\boldsymbol{\epsilon}\parallel_2^2\right]. 
\end{equation}

\vspace{0.1cm}
\noindent\textbf{$\bullet$ Modeling of the $\hat{\boldsymbol{\epsilon}}_{\boldsymbol{\theta}}$}. During the reactant generation stage, each atom contains both a feature vector and 3D coordinates. To preserve equivariance of $\hat{\boldsymbol{\epsilon}}_{\boldsymbol{\theta}}$ to coordinate rotations and translations (details about the equivariance are provided in Proposition 1), we utilize the EGNN~\cite{VictorEmiel21} to model $\hat{\boldsymbol{\epsilon}}_{\boldsymbol{\theta}}$. Define the feature of atom $i$ in the denoising time step $t$ as $\mathbf{z}_{i,t}=[\mathbf{z}_{i,t}^{(\rm x)},\mathbf{z}_{i,t}^{(\rm h)}]$ and the final reactant $\mathcal{R}$ containing $n$ atoms as $\mathcal{R}=\{v_{i}\}\vert_{i=1}^{n}=\{\mathcal{S},\mathcal{Q}\}$, where $\mathcal{S}=\{v_{i}\}\vert_{i=1}^{m}$ is a set of $m$ atoms in the synthon (obtained from the first stage), $\mathcal{Q}=\{v_{i}\}\vert_{i=m+1}^{n}$ is a set of $n-m$ atoms need to be generated. The process of determining the number of atoms is detailed in Section \ref{ssy0724:pred_num_atom} (supplementary material). Following the previous work \cite{hoogeboom2022equivariant}, $\hat{\boldsymbol{\epsilon}}_{\boldsymbol{\theta}}(\mathbf{z}_{i,t}, t)$ can be written as:
\begin{equation}
\fontsize{9.3pt}{12pt}\selectfont
    \hat{\boldsymbol{\epsilon}}_{\boldsymbol{\theta}}(\mathbf{z}_{i,t}, t) = \left[\mathbf{e}_{i,t}^{({\rm x}),L}, \mathbf{e}_{i,t}^{({\rm h}),L}\right]-\left[\mathbf{z}_{i,t}^{({\rm x})},\mathbf{0}\right],
\end{equation}
where $\left[\mathbf{e}_{i,t}^{({\rm x}),L}, \mathbf{e}_{i,t}^{({\rm h}),L}\right]$ is the embedding of atom $i$ output by an $L$-layer EGNN (time $t$), and its computation process is:
\begin{equation}
\fontsize{9.3pt}{12pt}\selectfont
\label{ssy0119:egnn}
    \begin{aligned}
        & \mathbf{m}_{ij} = \phi_e\left(\mathbf{e}_{i,t}^{({\rm h}),l-1},\mathbf{e}_{j,t}^{({\rm h}),l-1}, \parallel\mathbf{d}_{ij}^{l-1}\parallel_2^2\right), \\
        & \mathbf{e}_{i,t}^{({\rm h}),l}= \phi_{h}\Big(\mathbf{e}_{i,t}^{({\rm h}),l-1},\sum_{j\neq i}\mathbf{m}_{ij}\Big),\\
        & \mathbf{e}_{i,t}^{({\rm x}),l}= 
        \left\{
        \begin{aligned}
        & \mathbf{e}_{i,t}^{({\rm x}),l-1} + \sum_{i\neq j}\frac{{\mathbf{d}_{ij}^{l-1}}\phi_r\left(\mathbf{e}_{i,t}^{({\rm h}),l-1},\mathbf{e}_{j,t}^{({\rm h}),l-1}\right) }{1+\parallel\mathbf{d}_{ij}^{l-1}\parallel_2^2}, v_i \notin \mathcal{S},\\
        & \mathbf{e}_{i,t}^{({\rm x}),l-1}, v_i \in \mathcal{S},
        \end{aligned}
        \right.\\
    \end{aligned}
\end{equation}
with $\mathbf{d}_{ij}^{l-1}=\mathbf{e}_{i,t}^{({\rm x}),l-1}-\mathbf{e}_{j,t}^{({\rm x}),l-1}$, $\phi_e$/$\phi_h$/$\phi_r$ is a multi-layer perceptron. In the reactant generation, a node contains 3D coordinates $\mathbf{u}^{(\rm x)}$ and features $\mathbf{u}^{(\rm h)}$. Processing such features associated with 3D coordinates requires operations that respect the symmetry of the data, i.e., $\mathbf{u}^{(\rm h)}$ should be invariant to group transformations and $\mathbf{u}^{(\rm x)}$ should be related to rotations/translations. Details as shown in Proposition 1:

\vspace{0.1cm}
\noindent\textbf{Proposition 1} \emph{The update of node features in Eq. \ref{ssy0119:egnn} satisfies permutation invariance, while the update of coordinates changes with the input coordinate attributes. Formally, denote the update rules in Eq. \ref{ssy0119:egnn} as an abstract function $f(\cdot)$, then we have $f(\cdot)$ satisfies $\mathbf{z}^{(\rm x)}, \mathbf{z}^{(\rm h)}=f\left(\mathbf{u}^{(\rm x)}, \mathbf{u}^{(\rm h)}\right)$ and $\mathbf{U}\mathbf{z}^{(\rm x)}+\mathbf{t}, \mathbf{z}^{(\rm h)}=f\left(\mathbf{U}\mathbf{u}^{(\rm x)}+\mathbf{t}, \mathbf{u}^{(\rm h)}\right)$, where $\mathbf{U}$ is an orthogonal matrix to rotate $\mathbf{u}^{(\rm x)}$, and $\mathbf{t}$ is a translation vector.} 

\vspace{0.1cm}
\noindent\emph{Proof.} {See Section \ref{ssy0724:pp1} (supplementary material).}

It can be seen that we keep the coordinates of the atoms in the $\mathcal{S}$ unchanged (Eq. \ref{ssy0119:egnn}), so that the generation of reactants is conditioned on the synthon obtained in the first stage. Essentially, introducing conditions (defined as $\tilde{\mathbf{c}}$) is equivalent to shifting the mean \cite{DBLP:conf/icml/JaschaEric15} of the reverse sampling process, as follows: 

\vspace{0.1cm}
\noindent\textbf{Proposition 2} \emph{The benefit of introducing conditions $\mathbf{\tilde{c}}$ lies in injecting prior knowledge into the reverse sampling process, i.e., $p_{\boldsymbol{\theta}}(\mathbf{z}_{t-1}\vert\mathbf{z}_{t},\tilde{\mathbf{c}})=\mathcal{N}(\mathbf{z}_{t-1};\boldsymbol{\mu}_{\boldsymbol{\theta}}(\mathbf{z}_{t}, t)+\mathbf{R}_{\boldsymbol{\theta}}(\mathbf{z}_{t}, t)\nabla_{\mathbf{z}_{t}}\log p(\tilde{\mathbf{c}}\vert\mathbf{z}_{t}), \mathbf{R}_{\boldsymbol{\theta}}(\mathbf{z}_{t}, t))$.}

\vspace{0.1cm}
\noindent\emph{Proof.} See Section \ref{ssy0724:pp2} (supplementary material). 

The Proposition 2 indicates that the reverse sampling process with the mean $\boldsymbol{\mu}_{\boldsymbol{\theta}}(\mathbf{z}_{t}, t)+\mathbf{R}_{\boldsymbol{\theta}}(\mathbf{z}_{t}, t)\nabla_{\mathbf{z}_{t}}\log p(\tilde{\mathbf{c}}\vert\mathbf{z}_{t})$ and variance $\mathbf{R}_{\boldsymbol{\theta}}(\mathbf{z}_{t}, t)$, where the additional term $\mathbf{R}_{\boldsymbol{\theta}}(\mathbf{z}_{t}, t)\nabla_{\mathbf{z}_{t}}\log p(\tilde{\mathbf{c}}\vert\mathbf{z}_{t})$ guides the model to generate reactants that are more line with the prior (i.e., conditions).
The prior $\mathbf{R}_{\boldsymbol{\theta}}(\mathbf{z}_{t}, t)\nabla_{\mathbf{z}_{t}}\log p(\tilde{\mathbf{c}}\vert\mathbf{z}_{t})$ makes the generation of reactants more controllable (preventing deviation from chemical principles), thereby improving the final hit rate such as the top-1 accuracy.

\section{Experiments}
\begin{table*}
\small
  \centering
    {
    \begin{tabular}{c|c|p{1.1cm}<{\centering}p{1.1cm}<{\centering}p{1.1cm}<{\centering}|p{1.1cm}<{\centering}p{1.1cm}<{\centering}p{1.1cm}<{\centering}}
    \toprule[1.2pt]
    \multicolumn{2}{c|}{\multirow{2}[4]{*}{Baselines}} & \multicolumn{3}{c|}{Top-$k$ accuracy (unknown class)} & \multicolumn{3}{c}{Top-$k$ accuracy(known class)} \bigstrut\\
    \cline{3-8}    \multicolumn{2}{c|}{} & $k$=1   & $k$=3   & $k$=5   & $k$=1   & $k$=3   & $k$=5 \bigstrut\\
    \hline
    \multicolumn{1}{c|}{\multirow{6}[12]{*}{{\begin{tabular}[c]{@{}c@{}}Template-\\Based\end{tabular}}}} 
                         & MHNreact \cite{DBLP:journals/jcim/SeidlRenz22} & 51.8  & 74.6  & 81.2  & -     & -     & - \bigstrut\\
    \cline{2-8}          & GLN \cite{dai2019retrosynthesis}   & 52.5  & 69.0    & 75.6  & 64.2  & 79.1  & 85.2 \bigstrut\\
    \cline{2-8}          & LocalRetro \cite{chen2021deep} & 53.4  & \textbf{77.5} & \textbf{85.9} & 63.9  & \textbf{86.8}  & \textbf{92.4} \bigstrut\\
    \cline{2-8}          & GraphRetro \cite{somnath2021learning} & 53.7  & 68.3  & 72.2  & 63.9  & 81.5  & 85.2 \bigstrut\\
    \cline{2-8}          & RetroComposer \cite{DBLP:journals/bim/YanZhao22} & 54.5  & 77.2  & 83.2  & 65.9  & 85.8  & 89.5 \bigstrut\\
    \cline{2-8}          & Dual-TB \cite{sun2020energy} & \textbf{55.2} & 74.6  & 80.5  & \textbf{67.7}  & 84.8  & 88.9 \bigstrut\\
    \hline
    \multicolumn{1}{c|}{\multirow{8}[16]{*}{\begin{tabular}[c]{@{}c@{}}Template-\\Free\end{tabular}}} & Transformer \cite{vaswani2017attention} & 43.7  & 59.7  & 65.1  & -     & -     & - \bigstrut\\
    \cline{2-8}          & SCROP \cite{zhengrao20}  & 43.7  & 60.0    & 65.2  & 59.0    & 74.8  & 78.1 \bigstrut\\
    \cline{2-8}          & Transformer (\textit{Aug.}) \cite{tetko2020state} & 48.3  & -     & 73.4  & -     & -     & - \bigstrut\\
    \cline{2-8}          & RetroBridge  \cite{DBLP:conf/iclr/IgashovSchneuing24}  & 50.8  & \textbf{74.1}  & \textbf{80.6} & -     & -     & - \bigstrut\\
    \cline{2-8}          & GTA  \cite{seo2021gta}  & 51.1  & 67.6  & 74.8 & -     & -     & - \bigstrut\\
    \cline{2-8}          & Retroformer \cite{WanHsieh22} & 52.9  & 68.2  & 72.5  & 64.0    & \textbf{82.5}  & \textbf{86.7} \bigstrut\\
    \cline{2-8}          & Graph2SMILES \cite{tu2022permutation} & 52.9  & 66.5  & 70.0    & -     & -     & - \bigstrut\\
    \cline{2-8}          & Dual-TF \cite{sun2020energy}  & 53.6  & {70.7} & 74.6  & \textbf{65.7}  & 81.9  & 84.7 \bigstrut\\
    \cline{2-8}          & Chemformer \cite{irwin2022chemformer}  & \textbf{54.3} & -     & 62.3  & -     & -     & - \bigstrut\\
    \hline
    \multirow{6}[12]{*}{\begin{tabular}[c]{@{}c@{}}Semi-\\Template\end{tabular}} & MEGAN \cite{sacha2021molecule} & 48.1  & 70.7  & 78.4  & 60.7  & 82.0    & \textbf{87.5} \bigstrut\\
    \cline{2-8}          & G2Gs \cite{shi2020graph}  & 48.9  & 67.6  & 72.5  & 61.0    & 81.3  & 86 \bigstrut\\
    \cline{2-8}          & RetroXpert \cite{yan2020retroxpert} & 50.4  & 61.1  & 62.3  & 62.1  & 75.8  & 78.5 \bigstrut\\
    \cline{2-8}          & $\rm{G^2}$Retro \cite{DBLP:journals/cc/ChenAyinde23} & 51.4  & 72.1  & 78.2  & 59.4  & 79.4  & 84.2 \bigstrut\\
    \cline{2-8}          & RetroPrime \cite{DBLP:journals/cej/WangLi21} & 51.4  & 70.8  & 74.0    & 64.8  & 82.7  & 85 \bigstrut\\
    \cline{2-8}          & RetroDiff \cite{WangSong24} & 52.6  & 71.2  & 81.0    & -  & -  & - \bigstrut\\
    \cline{2-8}          & GDiffRetro (\textcolor{black}{Ours \ding{168}}) & \textcolor{black}{\textbf{58.9}} \ding{168} & \textcolor{black}{\textbf{79.1}} \ding{168} & \textcolor{black}{\textbf{81.9}} \ding{168} & \textcolor{black}{\textbf{67.6}} \ding{168} & \textcolor{black}{\textbf{84.1}} \ding{168} & \textcolor{black}{86.1} \ding{168} \bigstrut\\
    \bottomrule[1.2pt]
    \end{tabular}}
    \caption{Top-$k$ exact match accuracy (\%) on reaction dataset USPTO-50k. \textmd{The best result for each category is {bolded}. }}
  \label{table:main}%
\end{table*}

\begin{figure*}
    \centering
    \includegraphics[width=0.75\textwidth]{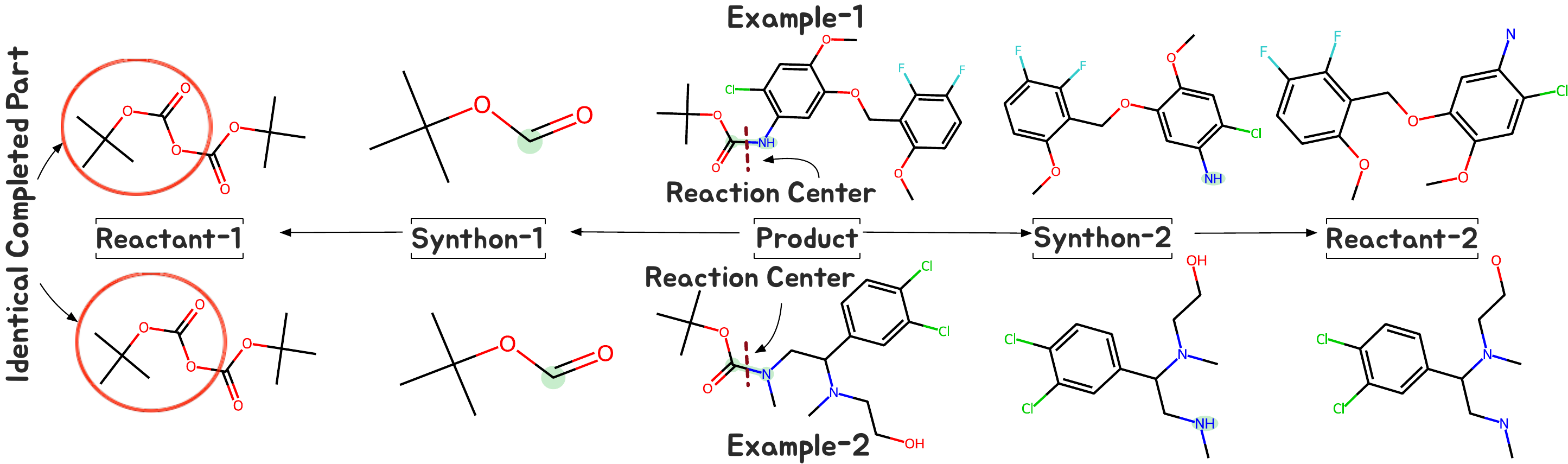}
    \caption{Depiction of the overall \textit{Retrosynthesis Prediction} process for examples in the \emph{protections} reaction class. \textmd{Reaction centers are highlighted on the products and synthons, while the completed parts are outlined with a circle on the reactants.}}
    \label{fig: end2end_generation}
\end{figure*}

\begin{figure*}
    \centering
    \includegraphics[width=0.73\textwidth]{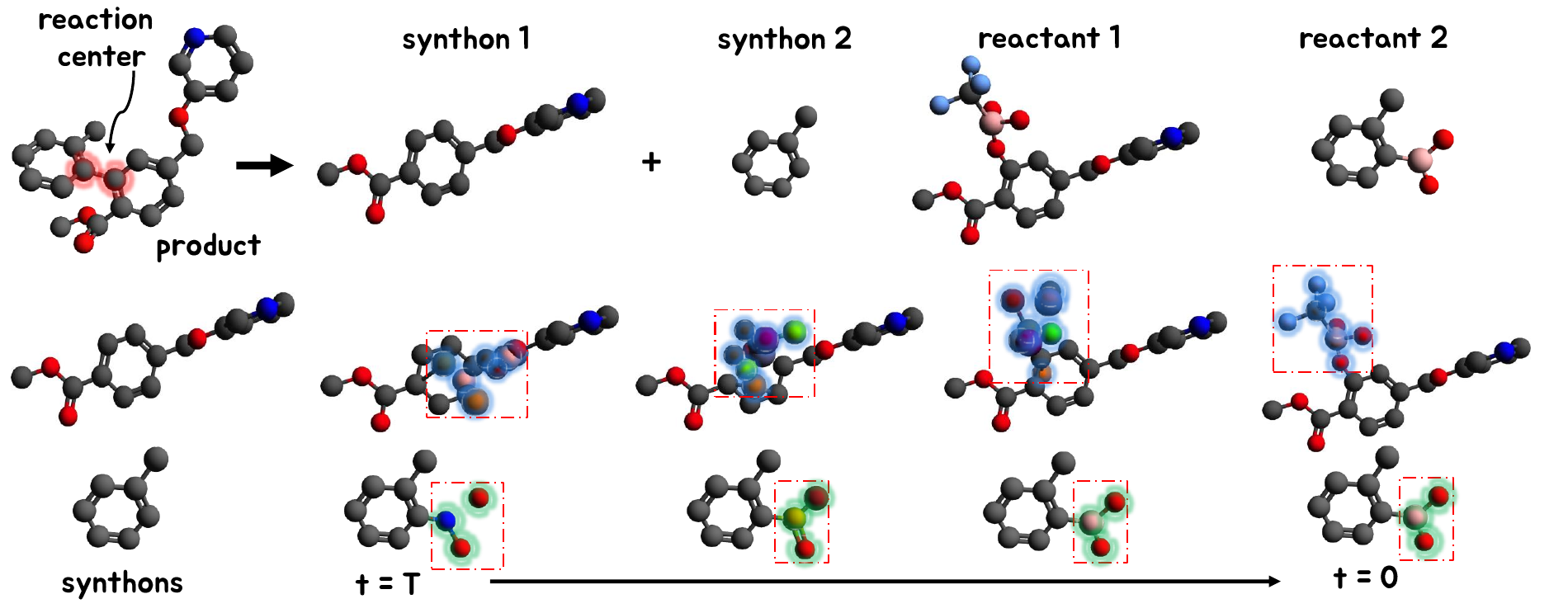}
    \caption{Trajectory of reactant generation. The atoms undergoing changes are highlighted with a rectangle.
    }
    \label{fig: case_study_3d}
\end{figure*}



\subsection{Experiment Setup}
We utilize the USPTO-50k dataset \cite{DBLP:journals/tec/Lowe2017}, detailed in Section B of the supplementary material, to assess the proposed method. The split of the dataset follows previous work \cite{liu2017retrosynthetic,shi2020graph}. Baselines are selected as follows: For \textit{Template-Based} methods, we select MHNreact \cite{DBLP:journals/jcim/SeidlRenz22}, GLN \cite{dai2019retrosynthesis}, LocalRetro \cite{chen2021deep}, GraphRetro \cite{somnath2021learning}, RetroComposer \cite{DBLP:journals/bim/YanZhao22}, and Dual-TB \cite{sun2020energy}. For \textit{Template-Free} methods, we select Transformer \cite{vaswani2017attention}, SCROP \cite{zhengrao20}, Retroformer \cite{WanHsieh22}, GTA \cite{seo2021gta}, Graph2SMILES (D-GCN) \cite{tu2022permutation}, Transformer (\textit{Aug.}) \cite{tetko2020state}, Dual-TF \cite{sun2020energy}, Chemformer \cite{irwin2022chemformer}, and RetroBridge  \cite{DBLP:conf/iclr/IgashovSchneuing24}. For \textit{Semi-Template} methods, we select MEGAN \cite{sacha2021molecule}, G2Gs \cite{shi2020graph}, RetroXpert \cite{yan2020retroxpert}, $\rm{G^2}$Retro \cite{DBLP:journals/cc/ChenAyinde23}, RetroPrime \cite{DBLP:journals/cej/WangLi21}, and RetroDiff \cite{WangSong24}. The definition of different types is given in Section \ref{ssy0119:introduction}. Following previous work \cite{liu2017retrosynthetic}, we employ the top-$k$ exact match accuracy as evaluation metric. More implementation details can be seen in Section \ref{ssy0724:imp_details} (supplementary material). Code available at https://github.com/sunshy-1/GDiffRetro.

\subsection{Performance Comparison} 
\label{ss:main result}
\textcolor{black}{As shown in Table \ref{table:main}, the top-1 result of GDiffRetro surpasses all template-free/semi-template based baselines, and most of the state-of-the-art template-based baselines. It's important to note that template-based methods rely heavily on external knowledge compared to template-free/semi-template based methods, making a direct comparison between template-based methods and template-free/semi-template based methods inherently unfair. To ensure fairness, we focus on performance gains within the category. Within the ``Semi-Template'' category, GDiffRetro achieves a relative improvement of 12.0\% (unknown class) and 4.3\% (known class, i.e. assuming the reaction class is known)
in terms of the top-1 metric compared to the second-best method.} This demonstrates that GDiffRetro can provide the most accurate retrosynthesis prediction with just a single attempt (this advantage is explained in the Proposition 2). 

\begin{table*}
\small
\centering
\captionsetup{singlelinecheck=false, justification=justified, width=0.9\textwidth}
  \centering
  \captionsetup[subtable]{labelformat=simple, labelsep=period}
  \begin{subtable}{0.45\textwidth} 
    \setlength{\tabcolsep}{2.25mm}
    \captionsetup{singlelinecheck=false, justification=justified, width=0.95\textwidth, justification=centerlast}
    {
    \begin{tabular}{c|ccccc}
    \toprule
    \multirow{2}[4]{*}{Setting ($Dual\text{-}\mathcal{G}$)} & \multicolumn{5}{c}{Top-$k$ Accuracy  \%} \\
    \cmidrule{2-6}            & $k$=1 & $k$=2 & $k$=3 & $k$=5 & $k$=10 \\
    \midrule
    $w/o$      & 81.4    & 93.4    & 96.5    & 98.6    & 99.6 \\
    $w$      & 86.2    & 95.1    & 97.4    & 98.8    & 99.6 \\
    \bottomrule
    \end{tabular}}
    \caption{Top-$k$ accuracy of \textit{Reaction Center Prediction}.}
    \label{subtab:ablation-stage1}
  \end{subtable}%
  \hspace{0.05\textwidth} 
  \begin{subtable}{0.45\textwidth} 
    \setlength{\tabcolsep}{3mm}
    \captionsetup{singlelinecheck=false, justification=justified, width=0.95\textwidth,justification=centerlast}
    {
    \begin{tabular}{c|cccc}
    \toprule
    \multirow{2}[4]{*}{Setting ($Dual\text{-}\mathcal{G}$)} & \multicolumn{4}{c}{Top-$k$ Accuracy \%} \\
    \cmidrule{2-5}            & $k$=1     & $k$=3     & $k$=5     & $k$=10 \\
    \midrule
    $w/o$     & 53.1    & 76.1    & 79.7      & 81.4 \\ 
    $w$     & 58.9    & 79.1      & 81.9    & 83.8 \\ 
    \bottomrule
    \end{tabular}}
    \caption{Top-$k$ accuracy of \textit{Retrosynthesis Prediction}.}
    \label{subtab:ablation-end2end}
  \end{subtable}%
  \caption{Results of single \textit{Reaction Center Prediction} and end-to-end \textit{Retrosynthesis Prediction} (unknown class). \textmd{$w$ $Dual\text{-}\mathcal{G}$ means the $Dual\text{-}\mathcal{G}$raphs considered. $w/o$ $Dual\text{-}\mathcal{G}$ means the vanilla RGCN without $Dual\text{-}\mathcal{G}$raphs.}}
  \label{tab:ablation}
\end{table*}

\begin{figure}
    \centering
    \includegraphics[width=1\linewidth]{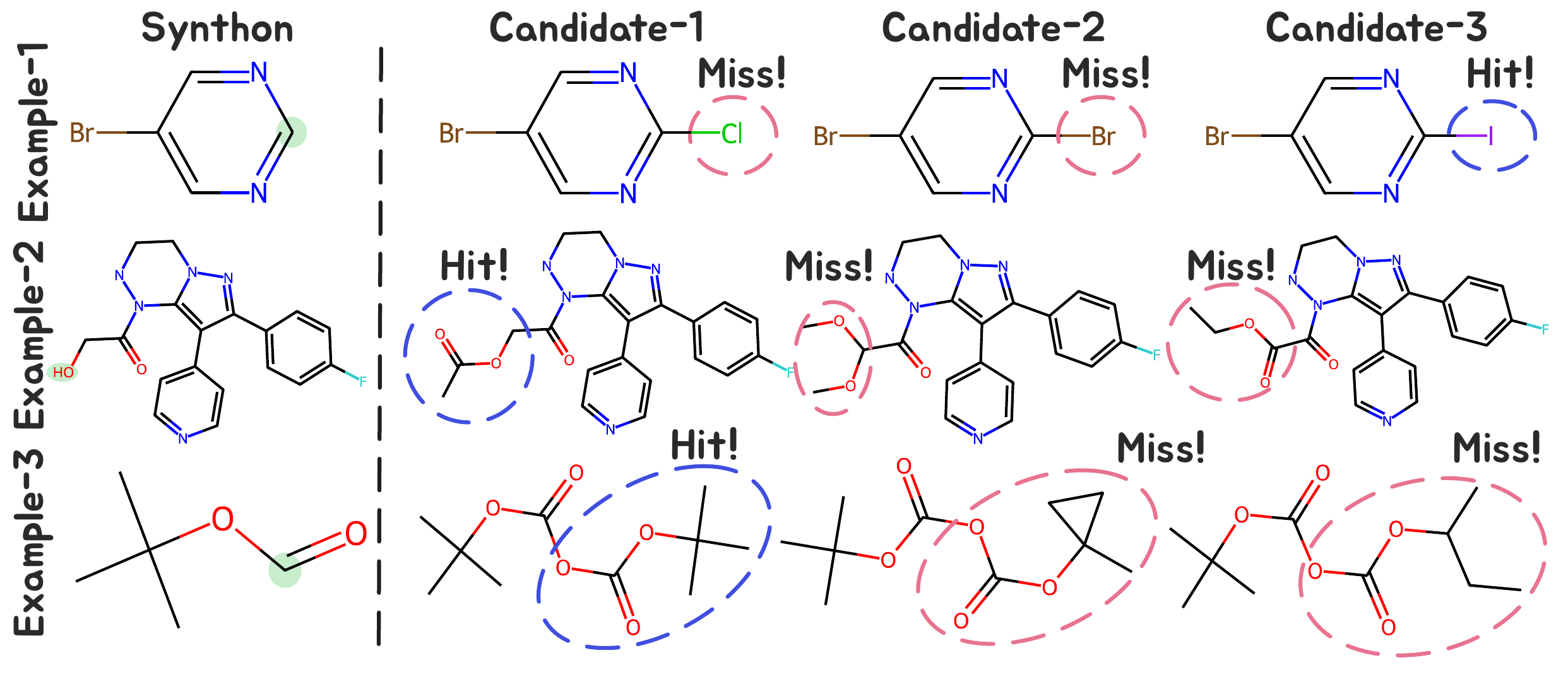}
        \captionsetup{belowskip=0pt}
        \caption{Top-3 generated reactants for distinct synthons. Correctly completed parts are denoted with \emph{Hit!}, while incorrect completions are marked with \emph{Miss!}.}
    \label{fig:top3}
\end{figure}

\begin{figure}[h]
    \centering
    \includegraphics[width=1\linewidth]{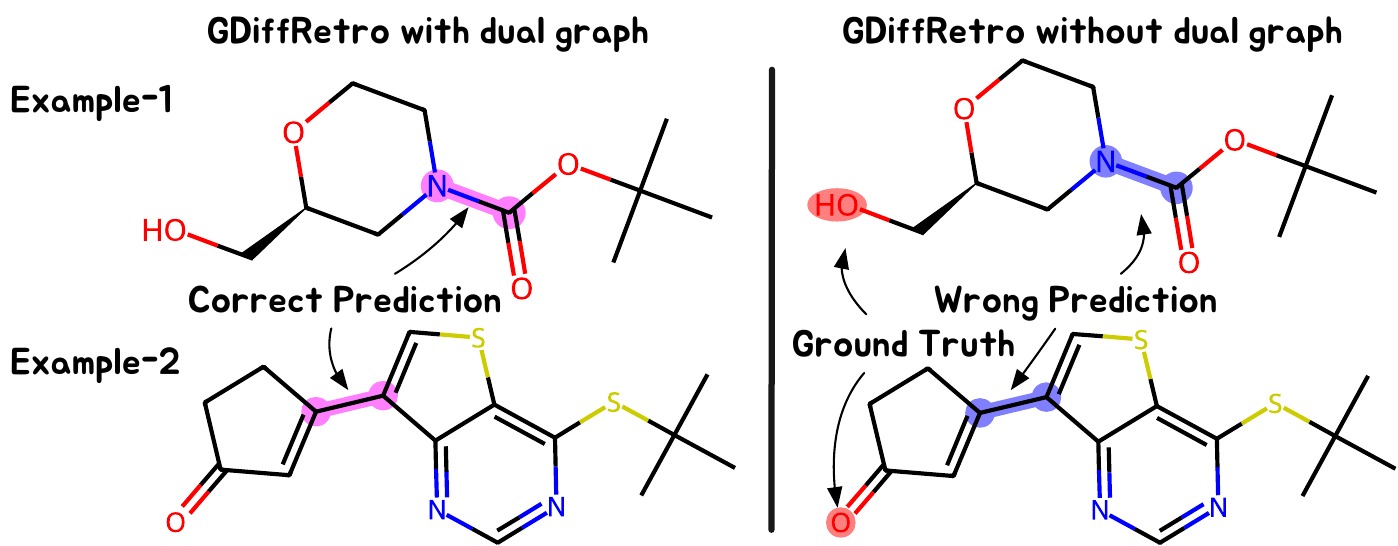}
        \caption{Illustration of reaction center on two products predicted by GDiffRetro, with dual graphs considered (left pair, the prediction and ground truth overlap) and without them (right pair, the prediction and ground truth are inconsistent).}
        \label{fig:stage1_ablation}
\end{figure}

In real-world applications, a relatively high single-attempt success rate (i.e., top-1 accuracy) is extremely important. This is because the reactants obtained a single retrosynthetic prediction usually not commercially available. Typically, we need to recursively conduct multiple retrosynthesis predictions to obtain the final synthesis route (akin to a search tree). Obviously, under this tree-like structure, a higher top-1 accuracy can greatly narrow the search space, thereby enhancing efficiency and reducing resource consumption. When examining the top-3 and top-5 metrics, GDiffRetro performs on par with all the template-free and semi-template baselines. Particularly, GDiffRetro almost outperforms all template-free/semi-template baselines in the top-3 and top-5 performances. It is worth noting that certain approaches, such as MEGAN, may exhibit significantly higher performance than GDiffRetro in terms of the top-5 result (known class). We attribute this difference to the limited number of sampling iterations in GDiffRetro, which may hinder its ability to generate an ample set of candidates. Mathematically, the diversity of results generated by a diffusion model is closely related to the sampling PDF's peakiness. In the conditional diffusion model, the peakiness depends on the mutual information between the condition and target. In our setting, the synthons (conditions) and reactants (targets) are similar, this leads to more concentrated sampling results, which in turn causes a decrease in the top-5 accuracy.

\subsection{Analysis and Visualization} \label{ss:case}
To assess the proficiency of GDiffRetro in learning reaction templates, we visualize the end-to-end retrosynthesis prediction for two examples from the same reaction class in Fig. \ref{fig: end2end_generation}. For two products belonging to the ``protections'' class, GDiffRetro accurately predicts the reaction centers. Then, it generates the reactants using two similar sets of synthons, formulating identical completed parts. These examples demonstrate that GDiffRetro's results are informed by its understanding of the reaction class, indicating its potential in capturing the underlying reaction template. The trajectory of reactant generation is provided in Fig. \ref{fig: case_study_3d}. Atoms to be generated start from a cluster of noise (with random coordinates and categories). As the denoising time step changes, reasonable atom types and coordinates gradually emerge.\\
\indent To demonstrate the validity and diversity of generated reactant candidates, we provide the top-3 results given by GDiffRetro on 3 different synthons in Fig. \ref{fig:top3}. It can be seen that GDiffRetro not only always matches in a one-shot trial, but also gives some reasonable results in the following sub-candidates. Among them, some examples generate all reactants with the same number of completed atoms, but GDiffRetro still provides diverse meaningful candidates.

\subsection{Ablation Study}
\label{ss:ablation}
To verify the effectiveness of the dual graphs we introduced to the RGCNs, we compare two sets of ablation experiments (\textbf{a.} top-$k$ accuracy of reaction center prediction, \textbf{b.} top-$k$ accuracy of end-to-end retrosynthesis prediction), with and without dual graphs introduced in the inference process. The results are shown in Table \ref{tab:ablation}a and Table \ref{tab:ablation}b, respectively. 

In Table \ref{tab:ablation}a, it is evident that GDiffRetro achieves a significant improvement in top-1 accuracy when incorporating dual graphs. GDiffRetro with $Dual\text{-}\mathcal{G}$ also consistently surpasses its counterpart without $Dual\text{-}\mathcal{G}$ among all other metrics. A clear correlation between the performance of the end-to-end retrosynthesis prediction and the $Dual\text{-}\mathcal{G}$ configuration can be observed in Table \ref{tab:ablation}b, following a similar trend to Table \ref{tab:ablation}a. GDiffRetro with $Dual\text{-}\mathcal{G}$ outperforms the version without it, showing approximately a 6\% increase in top-1 accuracy, and a slight improvement in top-3, top-5, and top-10 accuracy. The relatively modest improvement is attributed to the already high accuracy observed in these metrics.

The predictions for the reaction center of two test molecules, containing 1 and 3 rings, respectively, are depicted in Fig. \ref{fig:stage1_ablation}. With enhanced information provided by the dual graphs, GDiffRetro with $Dual\text{-}\mathcal{G}$ can offer more precise predictions. More examples from 10 distinct reaction classes are illustrated in Section \ref{ssy0724:more_examples} (supplementary material).

\section{Conclusion}
In this paper, we introduce GDiffRetro, a novel framework designed for retrosynthesis prediction. GDiffRetro notably incorporates a dual graph enhanced molecular representation for the reaction center identification, and introduces the 3D conditional diffusion model for reactant generation. Experimental results show that GDiffRetro not only surpasses current state-of-the-art models in top-1 accuracy, including those heavily reliant on templates, but also achieves competitive performance across top-3 and top-5 rankings. Through comprehensive ablation studies and detailed visualization, we confirm that the two key components proposed in GDiffRetro function independently and effectively.

\section*{Acknowledgments} This work was supported by the NSFC Young Scientists Fund (No. 9240127), the Donations for Research Projects (No. 9229129) of the City University of Hong Kong, the Early Career Scheme (No. CityU 21219323), and the General Research Fund (No. CityU 11220324) of the University Grants Committee (UGC).

\bibliography{aaai25}

\newpage
\clearpage
\appendix

\twocolumn[%
    \begin{center}
    \fontsize{15}{18}
    \textbf{GDiffRetro: Retrosynthesis Prediction with Dual Graph Enhanced\\ Molecular Representation and Diffusion Generation (Supplementary Material)}\\
    
    \end{center}
    \vspace{0.5cm}
]
\section{Notations}
\label{ssy0724:notations}
Vectors and matrices are denoted by bold lower case letters (e.g., $\mathbf{{a}}$) and bold upper case letters (e.g., $\mathbf{{A}}$), respectively. Calligraphic letters (e.g., $\mathcal{Q}$) denote sets, and $\vert \cdot \vert$ represents the number of elements in the set (e.g., $\vert\mathcal{Q}\vert$). Superscript $(\cdot)^{\top}$ stands for transpose. $\|$ denotes the concatenation operation. $\mathbb{R}^{m\times n}$ is real matrix space of dimension $m\times n$. $\mathbb{E}(\cdot)$ represents the statistical expectation. $\mathcal{N}(\boldsymbol{\mu}, \mathbf{R})$ denotes a Gaussian distribution with mean $\boldsymbol{\mu}$ and covariance matrix $\mathbf{R}$. $\mathcal{U}(a,b)$ denotes a uniform distribution with noise range from $a$ to $b$. $\mathbf{I}$ denotes an identity matrix. $\mathbb{I}(\cdot)$ is an indicator function. 

\section{Explanations of Chemical Terms}
\label{ssy0724:terms}
The chemical terms used in this paper and their corresponding explanations are summarized as follows:
\begin{enumerate}[left=2pt, label=$\bullet$]
    \item {\textbf{Retrosynthesis Prediction}. The retrosynthesis prediction problem is the inverse problem of the synthesis problem, that is, given a target molecule, we want to know which molecules it can be synthesized from.}
    \item {\textbf{Product}}. A product is a starting point of the retrosynthetic prediction, that is, the given target molecule.
    \item {\textbf{Reactant}}. Reactants are the output of the retrosynthesis prediction. Reactants can be combined through chemical reactions to synthesize the product.
    \item {\textbf{Synthon}}. A synthon is a subset of a reactant and the output of the reaction center identification. It is derived from the product by breaking the bond and may not be a legitimate molecule.
    \item {\textbf{Atom, Bond, Molecule $\Leftrightarrow$ Node, Edge, Graph}}. 
    In this paper, we model the retrosynthesis prediction as a graph mining task. Specifically, we treat molecules as graphs, where each atom in the molecule is represented as a node in the graph, and the chemical bonds between atoms are represented as different types of edges between nodes.
\end{enumerate}

\section{Dataset Details}
The USPTO-50k dataset is a standard single-step retrosynthesis benchmark. The USPTO-50k dataset is obtained from the open soruce patent database \cite{DBLP:journals/tec/Lowe2017}, which includes approximately 50,000 chemical reactions divided into 10 classes (details of the classes are shown in Table \ref{ssy0724:dataset_class_info}). It should be noted that chemical reactions containing multiple products are split into multiple single product reactions, with each reaction preserving the reactants from the original reaction. Reactions involving trivial products, such as inorganic ions and solvent molecules, are eliminated. Among the whole reaction dataset, 13.61\% of reaction centers are bonds within some ring, while 40.09\% have exactly one node on rings. Only 0.448\% of the graphs in the USPTO-50k and other molecular datasets are non-planar, making such cases relatively rare. Nevertheless, our model integrates both the original graph and a dual graph structure, enabling the GNN component to effectively handle non-planar graphs. Notably, all reported results are derived from the whole dataset, without excluding these special cases.

\begin{table}[h]
    \centering
    {
    \begin{tabular}{cc}
        \toprule[1.0pt]
        Class Name & \#Examples \\
        \midrule[1.0pt]
        Heteroatom alkylation and arylation & 15,204 \\
        Acylation and related processes & 11,972 \\
        Deprotections & 8,405 \\
        C-C bond formation & 5,667 \\
        Reductions & 4,642 \\
        Functional group interconversion (FGI) & 1,858 \\
        Heterocycle formation & 909 \\
        Oxidations & 822 \\
        Protections & 672 \\
        Functional group addition (FGA) & 231 \\
        \bottomrule[1.0pt]
    \end{tabular}}
    \caption{Classes in the USPTO-50k dataset.}
    \label{ssy0724:dataset_class_info}
\end{table}

\section{Proofs of Propositions}
Below, we offer proofs for several propositions that are not included in the main draft.
\subsection{Proof of Proposition 1}
\label{ssy0724:pp1}
Noting that $\mathbf{U}$ is an orthogonal matrix ($\mathbf{U}\mathbf{U}^{T}=\mathbf{I}$), we have  $\left\|\mathbf{U}\mathbf{x}\right\|^2=\mathbf{x}^{\top}\mathbf{U}^{\top}\mathbf{U}\mathbf{x}=\left\|\mathbf{x}\right\|^2$. Furthermore, we can deduce that the update of node features
\begin{equation}
\scalebox{0.8}{
    $
    \begin{aligned}
    &\phi_{h}\Bigg(\mathbf{e}_{i,t}^{({\rm h}),l-1},\sum\limits_{j\neq i}\phi_e\Big(\mathbf{e}_{i,t}^{({\rm h}),l-1},\mathbf{e}_{j,t}^{({\rm h}),l-1},\Big\|\mathbf{U}\mathbf{e}_{i,t}^{({\rm x}),l-1} + \mathbf{t} -\mathbf{U}\mathbf{e}_{j,t}^{({\rm x}),l-1}- \mathbf{t}\Big\|_2^2\Big)\Bigg)\\=& \phi_{h}\left(\mathbf{e}_{i,t}^{({\rm h}),l-1},\sum\limits_{j\neq i}\phi_e\left(\mathbf{e}_{i,t}^{({\rm h}),l-1},\mathbf{e}_{j,t}^{({\rm h}),l-1}, \left\|\mathbf{e}_{i,t}^{({\rm x}),l-1}-\mathbf{e}_{j,t}^{({\rm x}),l-1}\right\|_2^2\right)\right) 
    \end{aligned}
    $}
\end{equation}
is invariant to rotations and translations, and the update of the 3D coordinate attributes 
\begin{equation}
\scalebox{0.8}{
    $
    \begin{aligned}
    &\mathbf{U}\mathbf{e}_{i,t}^{({\rm x}),l-1} + \mathbf{t}+\mathbb{I}_{(v_i\in \mathcal{S})}\otimes\sum\limits_{i\neq j}{{\left(\mathbf{U}\mathbf{e}_{i,t}^{({\rm x}),l-1} + \mathbf{t} -\mathbf{U}\mathbf{e}_{j,t}^{({\rm x}),l-1}- \mathbf{t}\right)}}\\
    &~~~~~~~\times \phi_r\left(\mathbf{e}_{i,t}^{({\rm h}),l-1},\mathbf{e}_{j,t}^{({\rm h}),l-1}\right)\left({1+\left\|\mathbf{U}\mathbf{e}_{i,t}^{({\rm x}),l-1} + \mathbf{t} -\mathbf{U}\mathbf{e}_{j,t}^{({\rm x}),l-1}- \mathbf{t}\right\|_2^2}\right)^{-1}\\
    =&\mathbf{U}\Bigg[\mathbf{e}_{i,t}^{({\rm x}),l-1}+\mathbb{I}_{(v_i\in \mathcal{S})}\otimes\sum\limits_{i\neq j}\frac{{\left(\mathbf{e}_{i,t}^{({\rm x}),l-1} -\mathbf{e}_{j,t}^{({\rm x}),l-1}\right)}\phi_r\left(\mathbf{e}_{i,t}^{({\rm h}),l-1},\mathbf{e}_{j,t}^{({\rm h}),l-1}\right) }{1+\left\|\mathbf{e}_{i,t}^{({\rm x}),l-1}-\mathbf{e}_{j,t}^{({\rm x}),l-1}\right\|_2^2}\Bigg] + \mathbf{t}\\
    =&\mathbf{U}\mathbf{e}_{i,t}^{({\rm x}),l} + \mathbf{t}
    \end{aligned}
    $
}
\end{equation}
is consistent with the change in input coordinates, where $\mathbb{I}(\cdot)$ is an indicator function.
\subsection{Proof of Proposition 2}
\label{ssy0724:pp2}
According to the Bayes' theorem, we have
\begin{equation}
\scalebox{0.85}{
    $
    \begin{aligned}
         p_{\boldsymbol{\theta}}(\mathbf{z}_{t-1}\vert\mathbf{z}_{t},\tilde{\mathbf{c}})=p_{\boldsymbol{\theta}}(\mathbf{z}_{t-1}\vert\mathbf{z}_{t})\exp\Big[{\log p_{\boldsymbol{\theta}}(\tilde{\mathbf{c}}\vert\mathbf{z}_{t-1})-\log  p_{\boldsymbol{\theta}}(\tilde{\mathbf{c}}\vert\mathbf{z}_{t})}\Big].
    \end{aligned}
    $
    }
\end{equation}
Using a Taylor expansion around $\mathbf{z}_{t}$, the $\log p_{\boldsymbol{\theta}}(\tilde{\mathbf{c}}\vert\mathbf{z}_{t-1})$ can be further expressed as
\begin{equation}
\label{ssy0119:ctrl}
\scalebox{0.85}{
    $
    \begin{aligned}
        p_{\boldsymbol{\theta}}&(\mathbf{z}_{t-1}\vert\mathbf{z}_{t},\tilde{\mathbf{c}})\\
        \approx~&p_{\boldsymbol{\theta}}(\mathbf{z}_{t-1}\vert\mathbf{z}_{t})\exp\left\{{(\mathbf{z}_{t-1}-\mathbf{z}_{t})\nabla_{\mathbf{z}_t}\log p_{\boldsymbol{\theta}}(\tilde{\mathbf{c}}\vert\mathbf{z}_{t})+\frac{\partial}{\partial t}\log p_{\boldsymbol{\theta}}(\tilde{\mathbf{c}}\vert\mathbf{z}_{t})}\right\}\\
        \propto&\exp\Bigg\{{-\frac{1}{2}{\left[\mathbf{z}_{t-1}-\boldsymbol{\mu}_{\boldsymbol{\theta}}(\mathbf{z}_{t}, t))^{\top}{\mathbf{R}^{-1}_{\boldsymbol{\theta}}(\mathbf{z}_{t}, t)}(\mathbf{z}_{t-1}-\boldsymbol{\mu}_{\boldsymbol{\theta}}(\mathbf{z}_{t}, t)\right]}}\Bigg\}\\&\times\exp\Bigg\{{\mathbf{z}_{t-1}\nabla_{\mathbf{z}_t}\log p_{\boldsymbol{\theta}}(\tilde{\mathbf{c}}\vert\mathbf{z}_{t})}\Bigg\}\\
        \propto&\exp\Bigg\{{-\frac{1}{2}{\left[\mathbf{z}_{t-1}-\boldsymbol{\mu}_{\boldsymbol{\theta}}(\mathbf{z}_{t}, t))^{\top}{\mathbf{R}^{-1}_{\boldsymbol{\theta}}(\mathbf{z}_{t}, t)}(\mathbf{z}_{t-1}-\boldsymbol{\mu}_{\boldsymbol{\theta}}(\mathbf{z}_{t}, t)\right]}}\Bigg\}\\&\times \exp\Bigg\{{\left[\mathbf{z}_{t-1}-\boldsymbol{\mu}_{\boldsymbol{\theta}}(\mathbf{z}_{t}, t)\right]\nabla_{\mathbf{z}_t}\log p_{\boldsymbol{\theta}}(\tilde{\mathbf{c}}\vert\mathbf{z}_{t})}\Bigg\}\\
        \propto&\exp\Bigg\{{-\frac{1}{2}{\left[\mathbf{z}_{t-1}-\boldsymbol{\mu}_{\boldsymbol{\theta}}(\mathbf{z}_{t}, t)-\mathbf{R}_{\boldsymbol{\theta}}(\mathbf{z}_{t}, t)\nabla_{\mathbf{z}_{t}}\log p(\tilde{\mathbf{c}}\vert\mathbf{z}_{t})\right]^{\top}}}\\&\times{{{\mathbf{R}^{-1}_{\boldsymbol{\theta}}(\mathbf{z}_{t}, t)}\left[\mathbf{z}_{t-1}-\boldsymbol{\mu}_{\boldsymbol{\theta}}(\mathbf{z}_{t}, t)-\mathbf{R}_{\boldsymbol{\theta}}(\mathbf{z}_{t}, t)\nabla_{\mathbf{z}_{t}}\log p(\tilde{\mathbf{c}}\vert\mathbf{z}_{t})\right]}}\Bigg\}.
    \end{aligned}
    $
    }
\end{equation}

The Eq. \ref{ssy0119:ctrl} indicates that $\mathbf{z}_{t-1}$ follows the Gaussian distribution $\mathcal{N}(\boldsymbol{\mu}_{\boldsymbol{\theta}}(\mathbf{z}_{t}, t)+\mathbf{R}_{\boldsymbol{\theta}}(\mathbf{z}_{t}, t)\nabla_{\mathbf{z}_{t}}\log p(\tilde{\mathbf{c}}\vert\mathbf{z}_{t}), \mathbf{R}_{\boldsymbol{\theta}}(\mathbf{z}_{t}, t))$, given the state $\mathbf{z}_{t}$ and condition $\tilde{\mathbf{c}}$. 

\begin{figure}[h]
    \centering
    \includegraphics[width=0.8\linewidth]{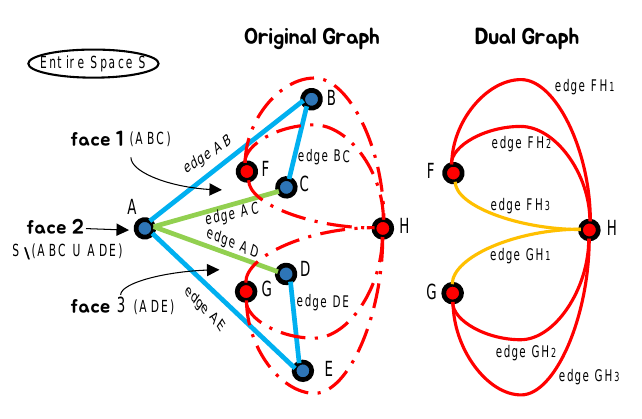}
    \caption{The concept of dual graphs.}
    \label{ssy0724:concept_dual_graph}
\end{figure}

\begin{figure}[h]
    \centering
    \includegraphics[width=0.8\linewidth]{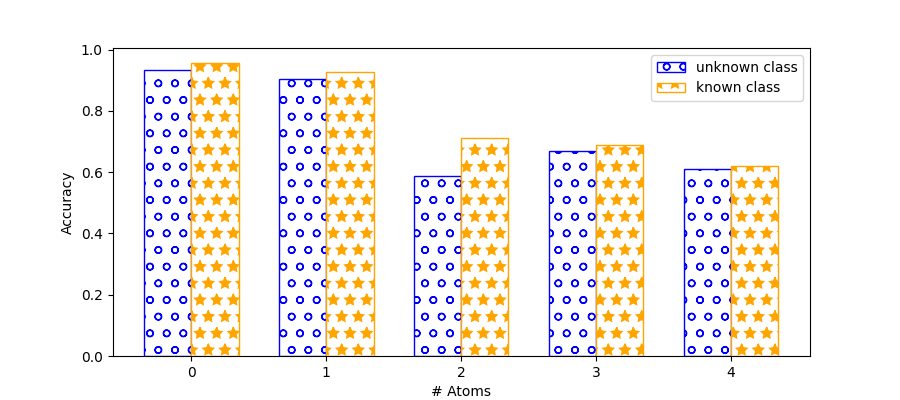}
    \includegraphics[width=0.8\linewidth]{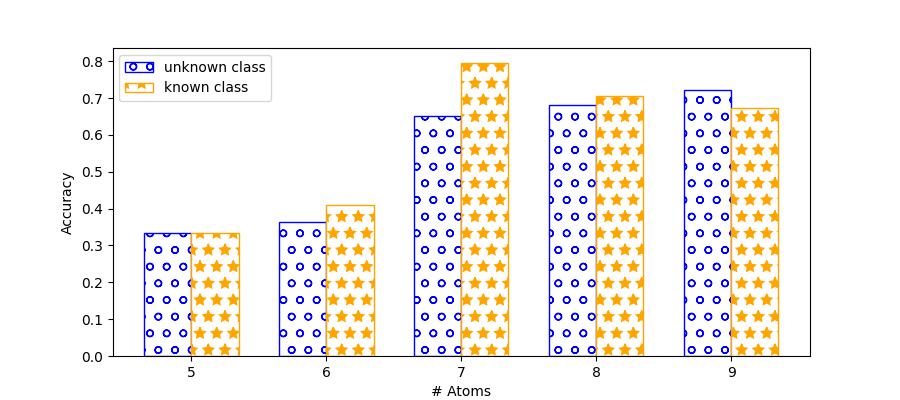}
    \caption{Accuracy of prediction of the number of atoms to be generated.}
    \label{fig:size_pred}
\end{figure}

\begin{figure*}[h]
    \centering
    \includegraphics[width=0.96\linewidth]{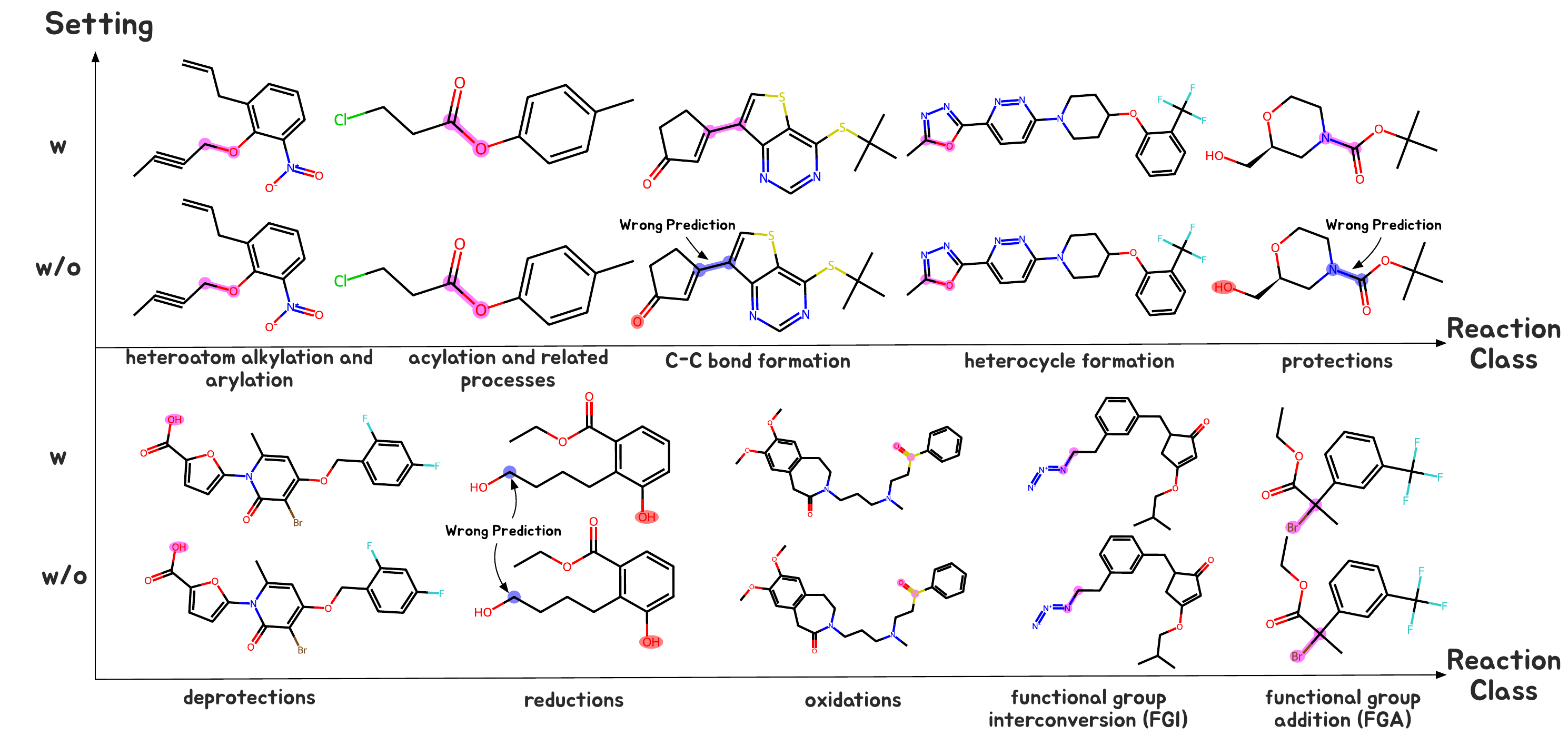}
    \caption{Illustration of reaction centers predicted by GDiffRetro (10 typical examples with 10 distinctive reaction classes), with dual graphs considered ($w$) and without them ($w/o$). The ground truths and predictions are highlighted in blue and red, respectively. Overlapping areas, indicating agreement, appear purple.}
    \label{fig:ablation_all}
\end{figure*}

\section{More Information about Dual Graphs}
\label{ssy0724:dual_graph_explain}
The dual graph is a concept in graph theory, primarily used for geometric and topological analysis of graphs. The dual graph describes a graph from the perspective of its faces (faces naturally exist in the graph and do not need to be predefined). As show in Fig. \ref{ssy0724:concept_dual_graph}, the original graph contains five nodes ($A$, $B$, $C$, $D$, and $E$) and six edges ($AB$, $AC$, $BC$, $AD$, $AE$, and $DE$). Additionally, the original graph divides the entire space $S$ into three regions, i.e, the face $ABC$, the face $ADE$, and the face $S \setminus (ABC \cup ADE)$. In the dual graph of the original graph, the three nodes $E$, $F$, and $G$ are located on the aforementioned three faces, respectively. The dual graph also contains six edges, i.e., $FH_{1}$, $FH_{2}$, $FH_{3}$, $GH_{1}$, $GH_{2}$, and $GH_{3}$.  The type of each edge in the dual graph depends on the
type of the edge it crosses in the original graph. For example, edge $FH_{1}$ and edge $FH_{2}$ each cross over edge $AB$ and edge $BC$ respectively, and since edges $AB$ and $BC$ in the original graph are of the same type (marked in blue), edges $FH_{1}$ and $FH_{2}$ in the dual graph are also of the same type (marked in red).\\
\indent In chemistry, faces and chemical properties are closely related. For example, rings in molecules can influence the stability of connected bonds through i) \textit{Strain Transmission}: Bond angles in rings deviate from ideal angles, resulting in strain within the ring. This strain can be transmitted to bonds connected to the ring, and change their bond lengths. ii) \textit{Electronic Inductive Effect}: Electron groups on the ring change the electron density on the connected bonds. Through the calculation of bond dissociation energies (BDE, which are proportional to stability) for the bonds {CCOC(=O)c1ccccc1} and O=C(OCc1ccco1)c1ccccc1, we observe that the introduction of the Tetrahydrofuran ring  results in a significant increase in the BDE of the adjacent C-C bond (from 90.1 kcal/mol to 107.5 kcal/mol). Therefore, modeling ``faces" in the first stage is reasonable, as they are closely related to bond stability.

\section{Implementation Details}
\label{ssy0724:imp_details}
We leverage the open-source RDKit library to construct molecular graphs based on SMILES. During the ``Reaction Center Identification'', we adopt the widely-used machine learning tool, \texttt{TorchDrug} \cite{zhu2022torchdrug}, to facilitate the training and evaluation processes. To obtain the 3D conformation of a molecular graph from SMILES, we adopt the data processing method employed by DeLinker \cite{imrie2020deep}, which involves comparing the conformations of a SMILES representation across all possibilities and selecting the one with the lowest energy. For the generation of SMILES representations of atomic point clouds produced by the diffusion model, we rely on OpenBabel \cite{o2011open}, an open-source tool in chemical research. To obtain the top-$k$ results, we sample 300 times during the inference process and select $k$ most frequent SMILES representations as the top-$k$ candidates. In the experimnets, we choose the top-$k$ exact match accuracy as evaluation metric. To facilitate meaningful comparisons, we consider different values of $k$ in our evaluations ($k=1,3,5$).

\section{Prediction of the Number of Atoms}
\label{ssy0724:pred_num_atom}
We treat the task of determining the number of atoms as a classification task \cite{igashov2022equivariant}, where classes are predefined according to the number of atoms to be generated. GDiffRetro first represents the synthon as a fully connected graph $\mathcal{G}^{*}$, where node features are one-hot encoded atom types and edge features are distances between nodes. Then,  an $L$-layer Graph Convolutional Network (GCN) is employed to predict the size of missing part of the synthon. After encoding through the GCN, embeddings of all nodes on the $\mathcal{G}^{*}$ are averaged and then softmaxed, ultimately yielding a probability distribution for predicting the number of atoms to be generated. The performance of the GCN adopted here is provided in Fig. \ref{fig:size_pred}. Specifically, the Fig. \ref{fig:size_pred} offers the accuarcy of prediction of the number of atoms to be generated $w$ and $w/o$ known reaction class as prior. It can be seen that knowing the type of reaction is helpful for predicting the number of atoms to be generated, as reaction types often imply fixed reaction templates. 

\section{More Visualization Examples}
\label{ssy0724:more_examples}
We select 10 typical examples with 10 distinctive reaction classes, and visualize their identified reaction center with dual graph ($w$) and without dual graph ($w/o$) configuration in Fig. \ref{fig:ablation_all}. It is easy to verify that dual graph enhanced molecular representations aids in the more accurate identification of reaction centers. For example, within the class ``C-C bond formation'', the model with the dual graph enhanced molecular representations correctly identifies the C-C bond as the reaction center, whereas its variant (the model without dual graph enhancement) incorrectly identifies the chemical bond around the oxygen atom as the reaction center.

\end{document}